\documentclass{article}

\PassOptionsToPackage{numbers, compress}{natbib}

\usepackage[preprint]{neurips_2021}

\usepackage[utf8]{inputenc} 
\usepackage[T1]{fontenc}    
\usepackage{hyperref}       
\usepackage{url}            
\usepackage{booktabs}       
\usepackage{amsfonts}       
\usepackage{nicefrac}       
\usepackage{microtype}      
\usepackage{xcolor}         
\usepackage{hyperref}

\usepackage{multicol}
\usepackage{wrapfig}
\usepackage{amsmath}
\usepackage{algorithm}
\usepackage[noend]{algpseudocode}
\usepackage{graphicx}
\usepackage{graphbox}
\usepackage{multirow}
\usepackage{xspace}
\usepackage{subfig}
\usepackage[switch]{lineno}  %
\let\oldhat\hat
\renewcommand{\vec}[1]{\mathbf{#1}}
\renewcommand{\hat}[1]{\oldhat{\mathbf{#1}}}
\newcommand{\algo}{CHIRL\xspace}
\newcommand{\context}[1]{{\small\textsc{#1}}\xspace}
\newcommand{\algcmt}[1]{{\textcolor{blue}{#1}}\xspace}
\newcommand{\ctxt}{\ensuremath{\kappa}\xspace}
\newcommand{\dset}{\ensuremath{\mathcal{D}}\xspace}
\newcommand{\dseq}{\ensuremath{\xi}\xspace}
\newcommand{\EVD}{\textsc{EVD}\xspace}
\newcommand{\feature}{\ensuremath{\mathbf{f}}\xspace}
\newcommand{\feat}{\feature}
\newcommand{\mdp}{\ensuremath{\mathcal{M}}\xspace}
\newcommand{\pol}{\ensuremath{\pi}\xspace}
\newcommand{\pmt}{\ensuremath{\theta}\xspace}
\newcommand{\pmtset}{\ensuremath{\boldsymbol\pmt}\xspace}

\newcommand{\cairlmh}{{HIRL}\xspace}
\newcommand{\cairlnsa}{{CHIRL-NoStateAbstraction}\xspace}
\newcommand{\nn}{\ensuremath{f}\xspace}

\newcommand{\TR}{Training Time\xspace}
\newcommand{\goalnav}{\mbox{\textsc{GoalNav}}\xspace}
\newcommand{\intnav}{\mbox{\textsc{JctNav}}\xspace}
\newcommand{\taxi}{\mbox{\textsc{Taxi}}\xspace}

\newcommand{\prob}{\ensuremath{p}\xspace}
\newcommand{\expt}{\ensuremath{\textrm{E}}\xspace}

\newcommand{\ie}{\textit{i.e.~}}
\newcommand{\eg}{\textit{e.g.}}
\newcommand{\etc}{\textit{etc.}}
\newcommand{\etal}{\textit{et~al.}\xspace}
\newcommand{\sss}{\scriptscriptstyle}

\usepackage{color}
\definecolor{fullred}{rgb}{0.95,.0,.1}
\newcounter{cmt}

\newenvironment{eq}
  {\vspace*{-2pt}\equation}{\vspace*{-2pt}\endequation}

\newcommand{\equref}[1]{Equation~\ref{#1}}
\newcommand{\figref}[1]{Figure~\ref{#1}}

\newcommand{\subfig}[1]{\textit{#1}}
\newcommand{\tabref}[1]{Table~\ref{#1}}

\graphicspath{{./figs/}}

\title{Context-Hierarchy Inverse Reinforcement Learning}

\author{%
   Wei Gao \\
    National University of Singapore\\
    \texttt{}
   \And
   David Hsu \\
  NUS\\
     \And
   Wee Sun Lee\\
  NUS\\
}

\begin{document}

\maketitle

\begin{abstract}
An inverse reinforcement learning (IRL) agent learns to act intelligently by observing expert demonstrations and learning the expert’s underlying reward function. Although learning the reward functions from demonstrations has achieved great success in various tasks, several other challenges are mostly ignored. Firstly, existing IRL methods try to learn the reward function from scratch without relying on any prior knowledge. Secondly, traditional IRL methods assume the reward functions are homogeneous across all the demonstrations. Some existing IRL methods managed to extend to the heterogeneous demonstrations. However, they still assume one hidden variable that affects the behavior and learns the underlying hidden variable together with the reward from demonstrations. They ignore all the causal dependencies over different factors. To solve these issues, we present \emph{Context-Hierarchy} IRL(CHIRL), a new IRL algorithm that exploits the context to scale up IRL and learn reward functions of complex behaviors. CHIRL models the context hierarchically as a directed acyclic graph; it represents the reward function as a corresponding modular deep neural network that associates each network module with a node of the context hierarchy. The context hierarchy and the modular reward representation enable data sharing across multiple contexts and context-specific state abstraction, significantly improving the learning performance. CHIRL has a natural connection with hierarchical task planning when the context hierarchy represents subtask decomposition. It enables to incorporate the prior knowledge of causal dependencies of subtasks and make it capable of solving large complex tasks by decoupling it into several subtasks and conquer each subtask to solve the original task. Experiments on benchmark tasks, including a large-scale autonomous driving task in the CARLA simulator, show promising results in scaling up IRL for tasks with complex reward functions.
\end{abstract}

\section{Introduction}
In inverse reinforcement learning (IRL), we model an agent's behaviors as a
Markov decision process (MDP) and learn from expert demonstrations a reward
function that induces desirable behaviors~\cite{NgRus00}. Most existing IRL methods focus on homogeneous demonstrations to infer one explicit reward function to explain the underlying behavior~\citep{ziebart2008maximum,wulfmeier2015maximum,finn2016guided}. Furthermore, in heterogeneous demonstration settings, the causal dependencies over each controlled factor are always ignored~\cite{krishnan2016hirl, yu2019meta}. How do we learn the reward function with heterogeneous demonstrations and also maintain the causal dependencies over different factors with \emph{limited data}?

\begin{wrapfigure}{r}{0.5\textwidth} 
  \vspace*{-6pt}
  \center
  \begin{tabular}{c@{\hspace{0.5in}}c}
    \includegraphics[height=.88in]{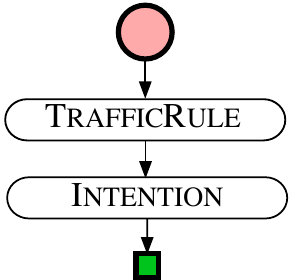}
    & \includegraphics[height=0.88in]{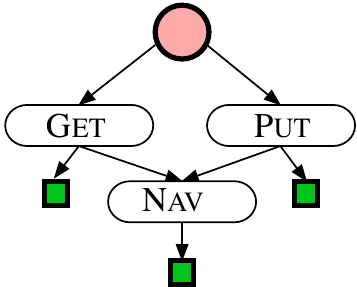}     \\
    (\subfig a) & (\subfig b)
  \end{tabular}
  \caption{Context hierarchy examples with DAG structure.  The root node is marked in red. Leaf nodes are
    marked in green.  All other nodes represent specific context variables. An edge represents dependency between context
    variables. 
    (\subfig a) ~An autonomous-driving
  task. (\subfig b)~Taxi ~\cite{dietterich2000hierarchical}. } \label{fig:example}
\end{wrapfigure}

Complex agent behaviors are often conditioned on the \emph{context}, and a
complex context can be decomposed and organized hierarchically.
\figref{fig:example} shows two illustrative examples.  The first one is a self-driving vehicle whose actions depend on the context: traffic rules and
passenger intentions.
 Can we pool data from different contexts---\eg,
driving on the right side and the left side of the road---and learn from
them simultaneously rather than learn in each context separately?  The second
example is the well-known taxi domain for hierarchical reinforcement
learning~\cite{dietterich2000hierarchical}.  A robot taxi navigates to pick up
a passenger and navigate to send the passenger to a destination. The context
represents the various subtasks forming the overall passenger delivery task.
Two subtasks, \context{Get} and \context{Put}, have the common subtask
\context{Nav}. The three subtasks have a causal dependency to finish the original task. Generally, different contexts may have commonalities and dependencies. How can we exploit the commonalities to share data and improve learning performance?

\emph{Context-Hierarchy IRL} (\algo) is a new IRL algorithm that uses a context hierarchy to decompose the reward function and represent it as a
modular deep neural network (NN).  Each node of the context hierarchy is
associated with a network module.  Each path, from the root of the hierarchy
to a leaf node, specifies a full context, and the network modules along the
path are composed to form the reward function for the context.  During the
learning process, data sharing across different contexts occur naturally
through the shared reward modules.  To increase sharing over different
contexts, \algo represents the context hierarchy as a directed acyclic graph
(DAG) instead of a tree. Such DAG structure naturally captures the causal dependency over different subtasks and enables to inject context dependency prior knowledge into IRL. To our best knowledge, this is the first work trying to explicitly make use of the prior knowledge of hierarchical tasks with causal dependencies in IRL domain.  

Task and motion planning are the principled approach when solving complex tasks~\cite{garrett2021integrated}. The original task are always decoupled into a sequence of "high-level" task planning and "low-level" motion planning, which provides better generalization and easy solution. Divide-and-conquer is prevalent in such task decomposition. How to properly decouple a complex task into subtasks are well-studied in the task and motion planning domain, for example, a very complex RoboCup~\cite{bai2012online}. In this paper, we choose one specific task decomposition method, MAXQ~\cite{dietterich2000hierarchical} to demonstrate how \algo benefits reward learning by injecting the prior task graph as domain knowledge. 


We evaluated \algo on several tasks, including the well-studied
taxi domain~\cite{dietterich2000hierarchical} and a large-scale autonomous driving task in
the CARLA simulator~\cite{dosovitskiy2017carla}, and obtained
promising results. The main contribution of this work is to make use of task decomposition to learn subtask reward to solve the original task by injecting the task graph into our reward learning while maintaining the task dependencies over subtasks. There are three main contributors to the improved
learning performance: \emph{data sharing} across contexts, 
context-specific \emph{state abstraction}, and \emph{subtask sharing}. 



\vspace*{-3pt}
\section{Context-Hierarchy Inverse Reinforcement Learning}
\vspace*{-2pt}
\subsection{Preliminaries}
\vspace*{-3pt}

\textbf{Maximum Entropy Inverse Reinforcement Learning (MaxEnt IRL)} In IRL, we model the agent behavior as an MDP 
$\mdp= (S, A, T, R)$, for state space $S$, action space~$A$,
state-transition function $T$, and reward function $R$. Given a state $s\in S$
and an action $a \in A$, the probabilistic state-transition function $T$
specifies the probability distribution of the resulting state~$s'$:
$T(s,a,s') = \prob(s' | s, a)$.
The state reward function assigns a scalar real value $R(s) $ for reaching
state $s$. It specifies the desired agent behavior. 
IRL assumes that the agent follows  an MDP policy $\pol\colon S\rightarrow A$,
which  prescribes an action $a=\pol(s)$ at every state $s\in S$  in order to maximize the
\emph{value}, \ie, the expected total reward over time:
$
V^\pol(s_0) = \expt\bigl[ R(s_0) + \gamma R(s_1)+ \gamma^2 R(s_2) + \cdots
  \bigr\vert \pol],
$
where  the expectation is taken over the distribution of state sequences $(s_0, s_1,
s_2, \ldots)$ resulting from executing \pol  with initial state $s_0$.
The constant  $\gamma\in [0,1)$ is a discount factor.
The standard IRL formulation~\cite{NgRus00} assumes that $S$, $A$, and $T$
are all known, and  we want to learn a reward function $R$ that induces an MDP
solution policy matching  the behavior and the performance of expert demonstrations as well
as possible. 




\textbf{MAXQ decomposition} The MAXQ decomposes a given MDP \textit{M} into a set of distinct sub-MDPs $\{M_0, M_1, \cdots, M_n\}$, arranged over a hierarchical structure, denoted as task graph. Specifically, $M_0$ is the root subtask (\ie solving $M_0$ solves the original MDP \textit{M}).

An \textit{normal} subtask $M_i$ is defined as a tuple $<T_i, A_i, R_i>$, where $i$ is the subtask index and

\begin{itemize}
\item{$T_i$ is the termination predicate that defines a set of active states, $S_i$, and a set of terminal states, $G_i$ for subtask $M_i$. The policy can only be executed on the active states, $S_i$. Whenever the agent reaches the $G_i$, subtask $M_i$ is terminated.}
\item{$A_i$ is a set of actions that can be performed to complete subtask $M_i$, which can either be primitive actions, or macro actions referring to other subtasks.}
\item{$R_i$ is the pseudo-reward for transitions from active states $S_i$ to terminal states $G_i$.}
\end{itemize}

A \textit{parameterized} subtask is defined as a tuple $<T_j, A_j, C_j, R_j>$, where $j$ is the group index and
\begin{itemize}
\item{$T_j, A_j$ is the shared transitions and actions.}
\item{$C_j$} is a set of parameters, denoted as context variable values.
\item{$R_j$} is the pseudo-reward for transitions from active states $S_j$ to terminal states $G_j$ given a context $C_j$
\end{itemize}

\subsection{Context-Hierarchy IRL}
In \algo, we create one reward function $R_\ctxt$ for each context \ctxt and
learn the set $\mathcal{R}$ of all context-hierarchy reward functions jointly. 
To do so, we augment a data set of trajectories with context labels.
Each element of a trajectory \dseq is a triplet of the form
$(\ctxt_t, s_t, a_t)$ for $t=0,1,2, \ldots$.  We preprocess the trajectories
so that each trajectory contains only tuples with the same context label, by
splitting a trajectory into multiple ones if necessary. We then break \dset
into subsets according to the context~\ctxt: $\dset = \bigcup_\ctxt \dset_\ctxt$.
To learn the context-hierarchy
rewards, we maximize the joint likelihood $\prob(\dset | \mathcal{R}) =
\prod_{\ctxt} \prob(\dset_\ctxt | R_\ctxt) $. 
We can directly derive the gradient of this objective function. Specifically, we use the negative log-likelihood of data $\mathcal{L_D}$ plus the $\ell_1$ and $\ell_2$ regularizers $\mathcal{L_{\pmtset}}$ over the reward function parameters as the loss function. 
\begin{align}
\label{eqn:obj}
\mathcal{L} = \mathcal{L_{D}} + \mathcal{L_{\pmtset}} = \sum_\kappa (\mathcal{L_{D_\kappa}} + \mathcal{L_{\pmtset_\kappa}}) 
= \sum_\kappa (- \log p(\mathcal{D}|R_\ctxt) + \lambda_{1}|\pmtset_\kappa| + \frac{1}{2}\lambda_{2}||\pmtset_\kappa||^{2} )
\end{align}
where $\lambda_{1}, \lambda_{2}$ are the coefficient of $l1, l2$ regularizers, $\pmtset_\kappa$ is the context parameters in the $R_\ctxt(\feat, \pmtset_\kappa)$ under the context $\kappa$.
We further derived the gradient of $\mathcal{L_D}$ w.r.t. reward function $R_\ctxt$ as (Detailed derivations can be found in the appendix): 
\begin{align}
\label{eqn:partial}
\frac{\partial \mathcal{L_D}}{\partial R_\kappa(\feat, \pmtset)} 
= \hat{\mu}_\vec{R_\kappa} -\hat{\mu}_\mathcal{D_\kappa} 
\end{align}
where $\hat{\mu}_\mathcal{D_\kappa}$ is the empirical state visitation count, 
and $\hat{\mu}_{\vec{R_\kappa}}$ is the expected state visitation count given current reward $R_\kappa$. 
Based on~\equref{eqn:obj},\ref{eqn:partial}, the total gradient can be written as:
\begin{align}
\frac{\partial \mathcal{L}}{\partial \pmtset}  = \frac{\partial \mathcal{L_D}}{\partial R_\ctxt(\feat, \pmtset_\kappa)} \frac{\partial R_\ctxt(\feat, \pmtset_\kappa)}{\partial \pmtset} + \frac{\partial \mathcal{L_{\pmtset}}}{\partial \pmtset} 
= (\hat{\mu}_{\vec{R_\ctxt}}- \hat{\mu}_{\mathcal{D}_\kappa} )\frac{\partial R_\ctxt(\feat, \pmtset_\kappa) }{\partial \pmtset} + \lambda_{1} sgn(\pmtset_\kappa) + \lambda_{2}\pmtset_\kappa
\end{align}

Our reward functions can be represented as neural network modules. 
It is straightforward to perform gradient-based optimization, using the backpropagation algorithm.
At the same time, it is natural to handle large state spaces with neural network reward representation. We define the reward as a function of state features
$\feat$ instead of~$s$, where $\feat \in \textbf{R}^d $ is an $d$-dimensional
state feature vector~\cite{NgRus00}, and then we can represent a reward function with neural network
$R(\feat; \pmtset)$, parametrized by a set  of parameters, \pmtset.

\begin{wrapfigure}{r}{0.5\textwidth} 
  \center
  \begin{tabular}{c@{\hspace{0.5in}}c}
    \includegraphics[height=0.8in]{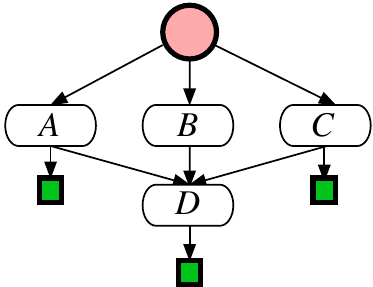}
    &\includegraphics[height=0.8in]{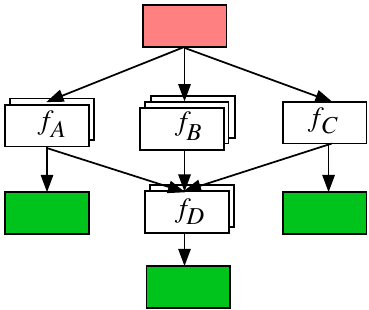}     \\
    (\subfig a) & (\subfig b)
  \end{tabular}
  \caption{A context hierarchy and the derived NN reward
    representation. (\subfig a) The root node is marked in red. Leaf nodes are
    marked in green.  All other nodes represent context variables. (\subfig b)
    Corresponding  context-hierarchy reward functions are represented as  interconnected
    NN modules, one for each node of the context hierarchy.}
\label{fig:cairl_definition}
\end{wrapfigure}

The context labels normally comes from the domain knowledge, it has a one-to-one mapping from the original task graph to the context hierarchy.
We represent a context as a set of factored context variables and organize them
into a DAG so that different contexts may share constituent variables. See
\figref{fig:cairl_definition}\subfig a for an example.  A node of the DAG represents a
context variable, and an edge between two nodes represents the dependency
between the corresponding context variables.
There are two special types
of nodes in the DAG: a dummy root node and dummy leaf nodes.  A
\emph{context path} goes from the root node to a leaf node.
It contains a sequence of
internal nodes with associated context variables that fully specify a context.

Given a context DAG, we construct  a NN representation of the context-hierarchy
reward function $R_\ctxt(\feat; \pmtset_\kappa)$.  For each node of the DAG, we
construct an NN module, which could be a multi-layer perceptron (MLP), a
convolutional neural network (CNN), \etc.  Let $V$ be the context variable
associated with an internal node. The corresponding NN module
$\nn_{\sss V}(\feat; \pmtset_v)$ is parametrized by a total of $|V|$ parameter
sets, one for each possible value $v$ of $V$.  For example, in
\figref{fig:cairl_definition}, the context variable $A$ may
represent the traffic rule and has two values: left-hand or right-hand driving.
So $f_A$ has two parameter sets. The dummy root node $V_0$
has no associated context variable.  The corresponding NN module
$\nn_0$ takes the state feature vector \feat directly as the input.
The dummy leaf nodes have no associated context variables, either. For
a leaf node $L$, $\nn_{\sss L}$ outputs a final reward value. 
If there is an edge from node
$V$ to node $V'$ in the context DAG, the output of $\nn_{\sss V}$
is then  connected to the input of $\nn_{\sss V'}$. 
Suppose that $(V_0, V_1, \ldots, V_{\sss N})$ is a context path, with root
node $V_0$  and leaf node $V_{\sss N}$ and that it specifies a context \ctxt with
$V_1=v_1, V_2=v_2, \ldots, V_{\sss N} = v_{\sss N}$.
We obtain the  reward function for \ctxt  by 
concatenating all the NN modules along the context path:
\begin{eq}
R_\ctxt(\feat; \pmtset_\kappa) =  f_{\sss N} ( \ldots  f_{1} (f_{0}(\feat; \pmtset_{v_0}) ;
\pmtset_{v_1}  ) \ldots  ; \pmtset_{v_{\sss N}}).   \label{eq:reward}
\end{eq}

The context hierarchy and the NN reward representation offer many benefits to
learning. They allow context-hierarchy reward functions to share NN modules across
contexts and drastically reduce the number of parameters to learn.
Consider again the example in \figref{fig:cairl_definition}. Suppose that
$|A| = 20, |B|= 30, |C| = 10, \text{and } |D| =20 $. There are
$|A| \times |B| \times |C| \times |D| = 1.2\times 10^5$ different contexts in
the worst case and thus $1.2\times 10^5$ parameter sets if we learn the reward
for each context independently. In contrast, \algo learns only $|A|+|C|+(|A|+|B|+|C|)|D| = 1250$
parameter sets, a reduction by two orders of magnitude,  by taking advantage of the factored NN reward representation, which comes from mostly the benefits of task decomposition.




\subsection{Algorithm}
\begin{algorithm}[h]
\caption{\algo}
\label{alg:cairl}
\begin{multicols}{2}
\small
\begin{algorithmic}[1]
\item[\algcmt{Compute the empirical state visitation}]
\item[\algcmt{frequency from demonstrations}]
\State {$\hat{\mu}_D, \hat{\nu}_D = svf(\gamma, T, D)$}
\item[\algcmt{Iterative model refinement}]
\For{\texttt{$i = 1:N$}}
\item[\algcmt{Compute the policy and }]
\item[\algcmt{its expected state visitation frequency}]
\State{Sample a context \ctxt}.
 \State{Extract $\hat{\mu}_{D_\kappa} \in \hat{\mu}_D \quad \quad \hat{\nu}_{D_\kappa} \in \hat{\nu}_D$. }
\State {Construct reward with NN modules $R_\ctxt(\feat,\pmtset_\kappa)$}
\State Get leaf node MDP dynamics $<S, A, T>$
\State {$\pi_\kappa = ValueIteration(R_\ctxt(\feat, \pmtset_\kappa), S, A, T, \gamma)$}
\State {$\mu_{R_\ctxt} = Propagate(\pi_\kappa, S, A, T, \hat{\nu}_{D_\kappa})$}
\item[\algcmt{Compute the negative log-likelihood loss and the graident.}] 
\State {$\mathcal{L} = \sum_\ctxt (- \log p(\mathcal{D}|R_\ctxt) + \lambda_{1}|\pmtset_\kappa| + \frac{1}{2}\lambda_{2}||\pmtset_\kappa||^{2})$}
\State{$\frac{\partial \mathcal{L}}{\partial \pmtset} = (\hat{\mu}_{\vec{R_\ctxt}}- \hat{\mu}_{\mathcal{D}_\kappa})
\frac{\partial}{\partial \pmtset}R_\ctxt(\feat, \pmtset_\kappa) + \lambda_{1} sgn(\pmtset_\kappa) + \lambda_{2}\pmtset_\kappa$}
\item[\algcmt{Back propagation to update reward parameters $\pmtset$}]
\State {nn\_backprop($\frac{\partial \mathcal{L}}{\partial \pmtset_\kappa}$)}
\EndFor
%
\end{algorithmic}

\bigbreak
\begin{algorithmic}[1]
\Procedure{svf}{$\gamma, T, D$}
\State {$E_{sa\kappa} = 0$}
\For {$\xi \in D$}
\For {$(s, \feat, a, \kappa) \in \xi$}
\State{$E_{sa\kappa} +=1$}
\EndFor
\EndFor
\State{$\mu_{D_{\kappa}}(s) = \sum_a {E_{sa\kappa}}$}
\State{$\nu_{D_\kappa} = \mu_{D_{\kappa}}$}
\For {$\xi \in D$}
\For {$(s, a, \kappa) \in \xi$}
\State{$\nu_{D_\kappa}(s') -= \gamma * T(s,a,s')$}
\EndFor
\EndFor
\State{$\mu_D = \bigcup_\kappa \mu_{D_\kappa}$}
\State{$\nu_D = \bigcup_\kappa \nu_{D_\kappa}$}
\State{\Return {$\mu_D, \nu_D$}}
\EndProcedure
\end{algorithmic}
\bigbreak

\begin{algorithmic}[1]
\Procedure{Propagate}{$\pi, S, A, T, \nu$}
\State {$E_0[\mu] = \nu$}
\For {$i = 0:N$}
\State{$E_{i+1}[\mu(s)] = \sum_{s',a} T(s,a,s')\pi(a|s)E_i[\mu(s')]$}
\EndFor
\State{\Return {$E_{N}[\mu]$}}
\EndProcedure
\end{algorithmic}
\end{multicols}
\end{algorithm}

The \algo algorithm with discounted factor $\gamma$ is described in~Algorithm\ref{alg:cairl}. 
The whole algorithm iteratively improves the learned reward in the outer loop and solves the optimal evaluation given the estimated reward in the inner loop. The gradient will guide the learned reward to minimize the state visitation difference of the induced policy and the expert demonstrations.

\section{Related Work}
Learning from demonstration \cite{argall2009survey} is a powerful approach to
intelligent agent behaviors.  Instead of learning a policy from the
demonstrated behaviors, IRL learns the underlying reward function, which
often leads to better generalization.
One key question of IRL is reward function representation.  Early work
represents the reward as a linear function of handcrafted features
\cite{abbeel2004apprenticeship,NgRus00,ziebart2008maximum}. Later work adopts
more powerful nonlinear function approximations, \eg, logical conjunction of
primitive features \cite{choi2013bayesian,LevPop10} or Gaussian processes
\cite{levine2011nonlinear}.  More recent work approximates the reward as a
deep NN and achieves state-of-the-art results on standard benchmark
tasks~\cite{finn2016guided,wulfmeier2015maximum}.
The above work,  however, treats the reward a single monolithic function with little structure.

Structured reward representations are beneficial.  Ziebart \etal propose to
learn a reward function parameterized by a \emph{context}
variable~\cite{ziebart2010modeling}, but no data is shared across the
contexts during learning.  Rothkopf and Ballard propose a modular reward
function representation as a set of weighted components for learning
visuomotor behaviors~\cite{rothkopf2013modular}.  All components jointly
contribute to the overall behavior according to the weights, which are
\emph{not} conditioned on the context.

Hierarchical decomposition is prevalent in planning and reinforcement learning
(RL) \cite{andre2002state,
  bai2012online,dietterich2000hierarchical,FraHo17,Kor85,Kno90,sutton1999between}.  It is, however, much less common in IRL. 
Krishnan \etal  treat a task as a Gaussian mixture model and try to
learn the subtask structure~\cite{krishnan2016hirl}, but there is no
hierarchical dependency among the subtasks. ~\cite{yu2019meta} tries to learn the meta reward functions based on a probabilistic context variable, however it doesn't consider any context dependencies. ~\cite{hwang2019option} and~\cite{venuto2020oirl} explored the possibility of extending IRL with temporally extended actions. They added the optional framework of semi-MDP into the IRL as the structured information for option reward to accelerate the learning tasks. However, these approaches are worked with only one-level hierarchy. In general, all the existing hierarchical approaches try to learn both the latent context and reward in the heterogeneous setting. However, we argue in this work, if we have prior domain knowledge, designing a framework to inject prior knowledge shares more benefits of data sharing, state abstraction and context causal dependency.


\section{Experiments}
\begin{wrapfigure}{r}{0.5\textwidth} 
\begin{minipage}[b]{0.18\linewidth}
\begin{tabular}{c}
\includegraphics[width=0.6in]{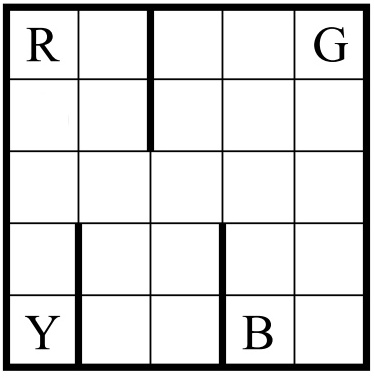} \\
(\subfig a) \\
\includegraphics[width=0.6in]{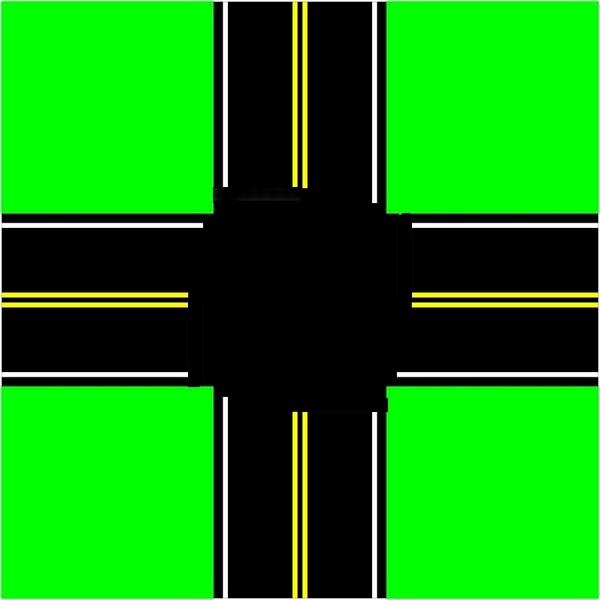} \\
(\subfig b) 
\end{tabular}
\end{minipage}
\hspace{0.04\textwidth}
\begin{minipage}[b]{0.68\linewidth}
\begin{tabular}{c}
\includegraphics*[width=1.9in]{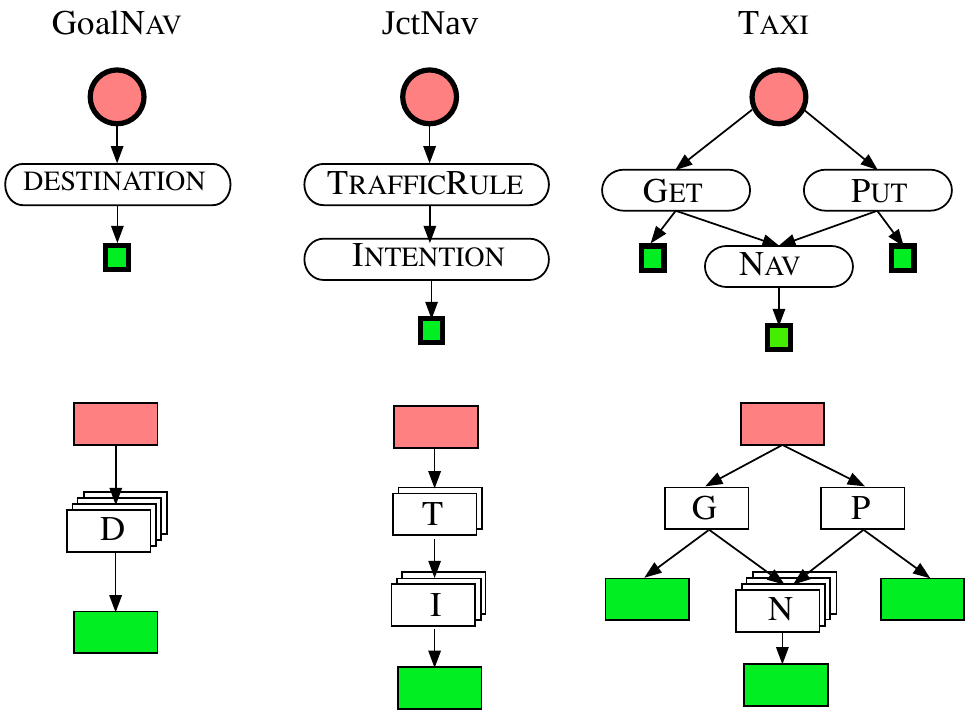}\\ 
{(\subfig c)} 
\end{tabular}

\end{minipage}
\caption{Evaluation tasks. (\subfig a) Navigation environment for \goalnav and \taxi. (\subfig b) Navigation environment~\intnav. (\subfig c)  Context hierarchies and corresponding reward networks for \goalnav, \intnav, and \taxi.}
\label{fig:benchmarks}
\end{wrapfigure}

We compared \algo with the state-of-the-art tabular IRL with neural network reward representations, DeepIRL~\cite{wulfmeier2015maximum}, the original MaxEnt~\cite{ziebart2008maximum} and HIRL~\cite{krishnan2016hirl}, on
benchmark tasks (\figref{fig:benchmarks}) of increasing difficulty to show the benefits of context-hierarchy. We also compared them in a large-scale autonomous driving task in the simulation. The results show that \algo consistently outperforms baselines. Note that there are other later works, AIRL~\cite{qureshi2018adversarial} and PEMIRL~\cite{yu2019meta} for continuous domain. We currently only consider tabular IRL for its simplicity to quantify all the metrics and the inner loop uses value iteration~\cite{chatterjee2008value} to have analytical solution. So we compare the corresponding best one in tabular versions. Extending to continuous domain is easy by changing the MDP to continuous version, which can be considered as an extension to this work.

\subsection{Experimental Setup}

MaxEnt, DeepIRL,  and \algo all apply the maximum entropy approach to IRL, but
differ in the reward function representation. For MaxEnt, the reward is a linear
function of state features. For DeepIRL, it is a nonlinear function
represented as a single monolithic neural network.  \algo
represents the reward as a network of NN modules, using the context hierarchy. \cairlmh is a variant of original HIRL, since original HIRL doesn't have the context label as input. Here, we add the context label as extra features to HIRL for a fair comparison.

We gave expert demonstration data, with the same data labeling,  to all algorithms for training.  Specifically, we appended context labels as part of the input data to MaxEnt, DeepIRL and HIRL. 
Finally, we performed a grid search to fine-tune the hyperparameters for all algorithms. All experiments are repeated $10$ times on a
computer server with an Intel Xeon CPU and a single GTX 2080 Ti GPU.


\subsection{Evaluation Metric}
Given an MDP\textbackslash{R},the goal of IRL is to learn the underlying R for the expert demonstrations. However, it is always hard to find a proper metric to measure the performance of IRL. \emph{Expected value difference} (EVD) ~\cite{levine2011nonlinear,wulfmeier2015maximum} provides a high-confidence upper bound on the policy loss incurred by using a evaluation policy $\pi_{R}$ under learned reward in place of optimal policy $\pi^*$, where $\pi^*$ is the optimal policy for the demonstration reward $R^*$.

The policy loss is captured via \textit{Expected Value Difference} (EVD) of $\pi_R$ under the true reward $R^*$, defined as 
$$EVD(R, R^*) = V^{\pi^*}_{R^*} - V_{R^*}^{\pi_R}$$


To better understand EVD, we first seek to bound the EVD. However, because the optimal policy $\pi^*$ is invariant to any negative scaling of the reward function, bounding EVD is ill-posed, as we can easily multiply the reward $R^*$ any $c > 0$ to scale EVD to anywhere in the range $[0, \infty)$. 
We seek to find a scale invariant EVD score across problems at different scales. We can rewrite the EVD as:
$$EVD(R, R^*) = cR^* (\mu^{\pi^*} - \mu^{\pi_R})$$
where $\mu^\pi$ is the expected feature counts over states or the occupancy measure of a policy $\pi$. 

To avoid the scaling issue, we can normalize the reward such that $c|R^*|_1 = 1$~\cite{syed2008game, pirotta2016inverse}. Note that this assumption only eliminates the trival all-zero reward function as a potential solution and all other reward can be appropriately normalized. All our benchmarks satisfy this assumption. 
The invariant normalized EVD is upper bounded by $EVD \leq \mu^{\pi^*} - \mu^{\pi_R}$ iff $c|R^*|_1 == 1$~\cite{brown2018efficient}.


The other important performance metric is the average training time, which measures the
average time for learning the reward function per epoch. Together, EVD and average training time indicate how well and how fast IRL algorithms learn.

\subsection{Comparison Results}
\subsubsection {{\goalnav}}
A robot navigates in a simple $5\times 5$ grid world with barriers
(\figref{fig:benchmarks}\subfig a).  Unlike standard IRL tasks, this
navigation task contains multiple destinations: R, G, B, and Y.
The robot has four actions, moving one cell in one of the four cardinal
directions, respectively. Each move succeeds with probability $0.8$ in the
desired direction and probability 0.1 in two neighboring directions.  The
ground-truth reward is $+5$ when the robot reaches the specified destination
and no reward elsewhere. A single context variable represents the
intended destination, and the reward depends on
it. 


\algo and \cairlmh   significantly outperform both
MaxEnt and DeepIRL in EVD (\figref{fig:env_quantitative}). DeepIRL does not perform well on this task. This performance difference demonstrates the benefits of context modeling with heterogeneous demonstrations. 
\goalnav has a very simple context hierarchy with only one level. The reward function representations for \algo and \cairlmh thus have almost the same architecture, resulting in a similar performance.


\subsubsection {{\intnav}}
This task models a simplified, but still a realistic autonomous driving scenario
at a cross junction.  The world is a $32\times 32$ grid, with two intersecting
roads placed randomly (\figref{fig:benchmarks}\subfig b).  A road consists of eight
lanes, four in each direction, with a center divider separating the two
directions.  The robot vehicle has two \emph{partial} state observations. One
provides coarse localization with 9 possible values: the vehicle is located
either in one of the eight lanes or anywhere else.  The other provides road
information in the local $3 \times 3$ neighborhood of the vehicle's current
location, indicating whether it is on-road, off-road, or on the center
divider.  The vehicle has nine actions, staying in place or moving by one cell
in the cardinal or diagonal directions.  Its behavior depends on two context
variables (\figref{fig:benchmarks}). One represents the traffic rule: drive on the
right side (RS) of or left side (LS) of the road.  The other represents the driving
intention: go straight (ST), turn left (TL), or turn right (TR).  The vehicle's
actions and thus the reward depend on these two variables.  The
ground-truth reward is $0$ if the vehicle drives in the correct lane according
to the traffic rule and $-10$ otherwise, as driving against the traffic is
extremely dangerous.  The reward is $-5$, if the agent mounts the center
divider. The reward is $+1$, if the agent successfully passes through the
intersection.

\begin{wrapfigure}{r}{0.5\textwidth} 
\vspace{-18pt}
\captionsetup[subfigure]{font=scriptsize,labelfont=scriptsize,position=top,labelformat=empty}
  \raisebox{-4\normalbaselineskip}[0pt][0pt]{\rotatebox[origin=c]{0}{\hspace{-5pt} \tiny \goalnav \hspace{-6pt} \;}}
\subfloat[EVD]{\includegraphics[width=0.21\columnwidth]{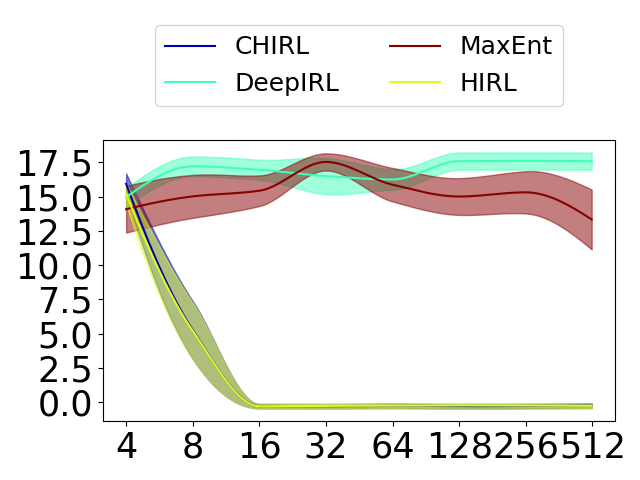}}
\hfill
\subfloat[Training Time]{\includegraphics[width=0.21\columnwidth]{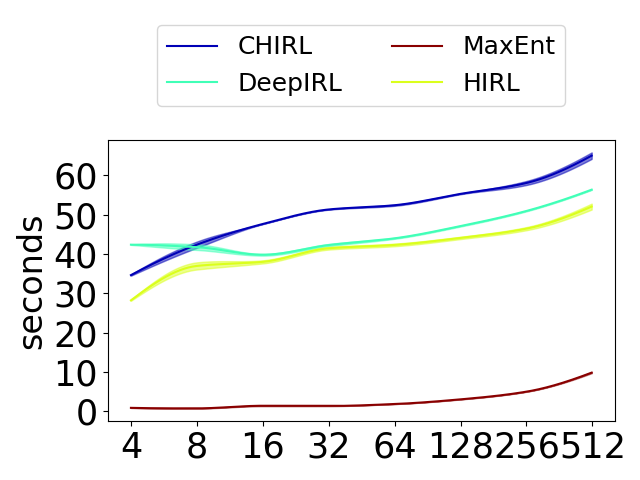}}

  \raisebox{-4\normalbaselineskip}[0pt][0pt]{\rotatebox[origin=c]{0}{\hspace{-5pt} \tiny \intnav  \;}}
\subfloat{\includegraphics[width=0.21\columnwidth]{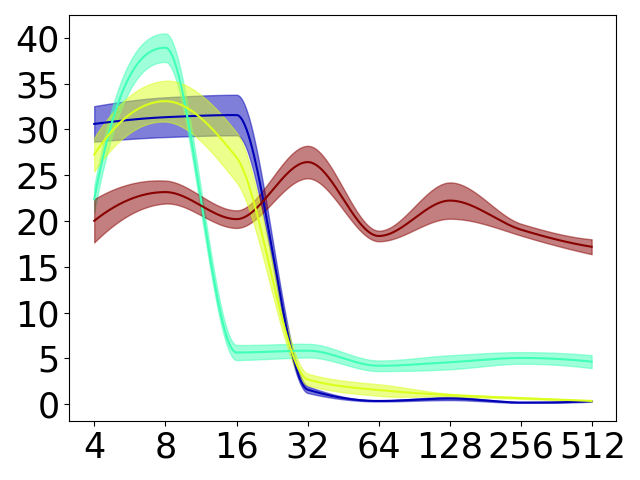}}
\hfill
\subfloat{\includegraphics[width=0.21\columnwidth]{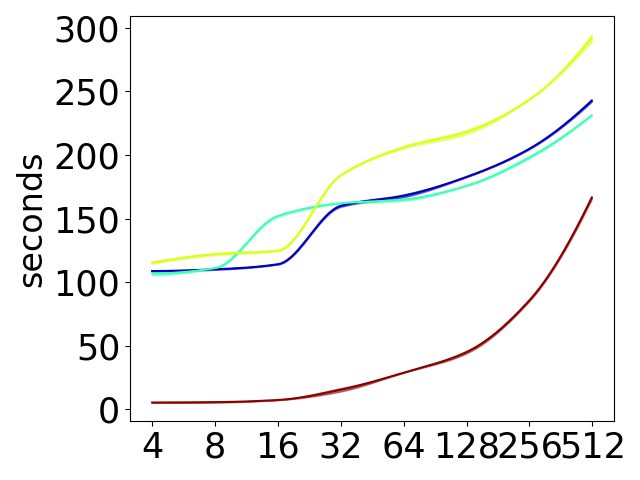}}

  \raisebox{-4\normalbaselineskip}[0pt][0pt]{\rotatebox[origin=c]{0}{\hspace{-5pt} \tiny \taxi \hspace{5pt} \;}}
\subfloat{\includegraphics[width=0.21\columnwidth]{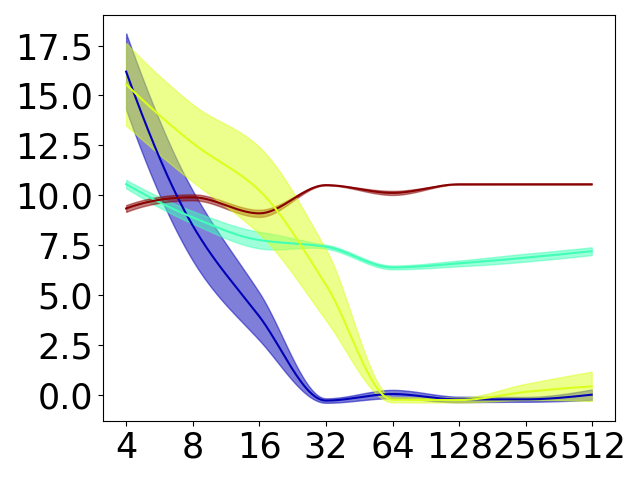}}
\hfill
\subfloat{\includegraphics[width=0.21\columnwidth]{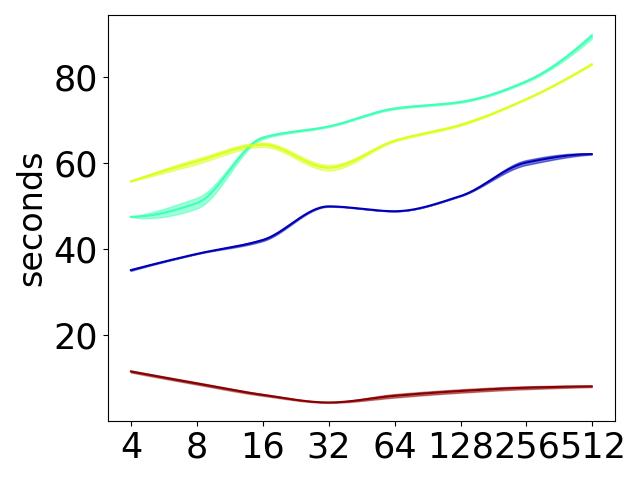}}
\caption{Performance comparison of \algo, MaxEnt, DeepIRL, and \cairlmh. The solid line represents the mean and shaded area represents the standard deviation. For EVD score, the lower the better.
If it reaches zeros, the induced policy from learn reward functions has the same distribution of expert policies. 
  In each plot, the horizontal axis represents the number of trajectories in the training data set \dset on the logarithmic scale.}
\label{fig:env_quantitative}
\vspace{-30pt}
\end{wrapfigure}

Again, \algo outperforms MaxEnt and DeepIRL in EVD
(\figref{fig:env_quantitative}).  
For \algo, the learned reward
matches the ground truth closely. For DeepIRL,  the learned reward
differentiates left-side and right-side driving, but without
context-hierarchy, it mixes up going straight and turns. This is, of course, highly desirable for junction navigation.
MaxEnt does not perform well in this task at all. Finally, compared with \goalnav, 
\intnav has a slightly more complex, two-level hierarchy. The advantage of \algo over \cairlmh starts to show. Notice that till now,
the causal dependency can be ignored in these two cases. 


\subsubsection{{\taxi}}

This task is originally designed for hierarchical MDP
planning~\cite{dietterich2000hierarchical}.  A taxi operates in the same grid
world (\figref{fig:example}\subfig b) as that for \goalnav, but has a different
task. There are four taxi terminals: R, G, B, and Y.  The taxi must get a
passenger at one terminal and send the passenger to a designated destination
terminal.
The state is a four-tuple $(x,y, p, d)$, where $(x, y)$ is the taxi location,
$p$ is the passenger location, $d$ is the destination terminal. The passenger
is located either in the taxi or one of the terminals. 
In each task instance, the initial taxi location, the initial passenger
location, and the destination terminal are chosen randomly.   
There are six primitive actions. The four navigation actions are the same as
those for \goalnav.  The two extra actions pick up or put down the passenger,
respectively.  Each action incurs $-1$ as the cost.  If the taxi
attempts to get or put the passenger at the wrong terminal or in a
non-terminal location, there is a penalty of $-10$.  The taxi gets a reward of
$20$ when it successfully delivers the passenger to the destination.

There are three leaf nodes in the TaxiDomain benchmark. In the $Get$ node, the state space contains $5 \times 5 \times 5 = 125$ states without considering the passenger's destination. There are four macro actions, $NavR, NavG, NavY, NavB$, and one primitive action $Pickup$. The demonstrations of $Get$ comes from all the data with $Pickup$ action and subtask $Nav(t)$. However, at this level, the macro action $Nav(t)$ can be directly finished by defining the termination predicate to move the taxi to the destination terminal. For example, $R$ terminal is at $(0,0)$. Any position in the grid world $s = (x, y)$, we can have that $T(s, Nav(R)) = (0, 0)$. It means that any position in the grid world can be moved to $R$ terminal after call macro action $Nav(R)$. We just record the starting position and termination position as the demonstration for the $Nav(t)$ macro actions.




For this task, the context hierarchy contains three variables---\textsc{Get}, \textsc{Put}, and \textsc{Nav}---each representing a subtask.
By leveraging the context hierarchy and the MAXQ decomposition,
\algo outperforms both DeepIRL and MaxEnt
(\figref{fig:env_quantitative}). Interestingly, \algo is much faster than DeepIRL in training time, because it represents \taxi as an HMDP with state abstraction rather than a 'flat' MDP. Also, with this complex subtask hierarchy with a total of 10 contexts, the advantage of \algo over \cairlmh becomes evident in both EVD and training time.


\subsection{Large-Scale Cost Function Learning for Autonomous Driving}


We extend \algo to learn the large-scale cost function for autonomous driving in the
high-fidelity CARLA simulator.
First, we extend the reward neural network to take an RGB image directly
as input. The image provides a top-down view of the local environment at the
vehicle's current location.  
Top-down views of the area surrounding
a vehicle are already widely available in commercial vehicles; these are
typically constructed using a multi-camera system on the vehicle~\cite{rathi2019multi}. 
Second, we extend the cost map from one specific scenario, i.e. all the benchmarks are one fixed environment, to anywhere egocentric cost map in various simulated towns. The reason that we can learn large scale cost map for autonomous driving is that all the driving behaviors can be treated as repeated local skills given the context information. Human drivers can drive anywhere once they learned enough basic local skills with traffic rule guidance. We aim to replicate the same human behaviors in \algo, since \algo is designed to be easily integrating traffic rules as the context information.    

\begin{table}
\vspace{-15pt}
\centering
\caption{
  Performance comparison for autonomous driving in the CARLA simulator. 
  Quantitative evaluation for driving in carla simulator with the negative
  log-likelihood (NLL) and modified Hausdorff distance (MHD) metrics in two
  scenarios: \textit{Full routes} and \textit{Intersections}. The \textit{Full
    routes} means the various full length paths from starting positions to the
  destinations. \textit{Intersections} means only the intersection areas are
  selected. Lower numbers represent models that are approximating human
  behavior with higher precision. We report the mean and standard deviation for each metric. Note that NLL is in logarithmic scale.
}

\label{tab:table1}
\scriptsize
\begin{tabular}{|c|c|c|c|}
\hline
Scenarios & Method & NLL & MHD (meters)\\ \hline
\multirow{3}{*}{Full routes} & DeepIRL & 12.548 (0.004) & 0.713 (0.022)  \\ \cline{2-4} 
& \cairlmh & 12.332 (0.005) & 0.542 (0.013) \\ \cline{2-4} 
  & \algo & \textbf{12.115 (0.006)} & \textbf{0.440 (0.011)} \\ \hline
\multicolumn{4}{|c|}{} \\ \hline
\multirow{3}{*}{Intersections} & DeepIRL & 12.809 (0.015) & 2.248 (0.103) \\ \cline{2-4} 
& \cairlmh & 11.998 (0.009) & 0.434 (0.012) \\ \cline{2-4} 
  & \algo & \textbf{11.808 (0.012)} & \textbf{0.234 (0.009)} \\ \hline
\end{tabular}
\vspace{-20px}
\end{table}

EVD is no longer valid for real-world applications because we don't have the ground truth reward for such applications. To measure the performance, we use similar metrics as done in~\cite{wulfmeier2017large}: the negative log-likelihood of the demonstration data (NLL) and the modified Hausdorff distance (MHD). NLL is used to measure the likelihood of expert demonstrations based on the reconstructed reward. MHD is a spatial metric for the similarity between the agent and demonstration trajectories. Since the original MaxEnt is always worse than DeepIRL when there is nonlinearity between the image and the reward function, we ignore the MaxEnt in the large-scale application. 


\tabref{tab:table1} shows the quantitative results of running different methods in the Carla simulator. Compared with the deepIRL baseline, \algo is better than the baseline in both NLL and MHD. Note that the NLL is measured in the log scale, the actual loss difference is large. \algo has lower MHDs in both scenarios, showing that we can learn the cost map well considering the different contexts. The intersection scenario has more performance gains, showing that the intersection areas are the critical places where the contexts matter the most.
It is obvious that when there are junctions, deepIRL cannot differentiate well. Furthermore, \algo works better than \cairlmh, which indicates the benefits of data sharing over the DAG structure. 

\subsection{Ablation Results} \label{sec:ablation}
To better understand the main contributors to CAIRL's performance, we performed an ablation study on \taxi, which has all features: multi-level data sharing, subtask state abstraction, and subtask sharing.

\subsubsection{Data Sharing}
\begin{wrapfigure}{r}{0.5\textwidth} 
\vspace{-50pt}
\captionsetup[subfigure]{font=scriptsize,labelfont=scriptsize,position=top,labelformat=empty}
  \raisebox{-3\normalbaselineskip}[0pt][0pt]{\rotatebox[origin=c]{0}{\hspace{1pt} \footnotesize (a)\; }}
\subfloat[\EVD]{\includegraphics[width=0.21\columnwidth]{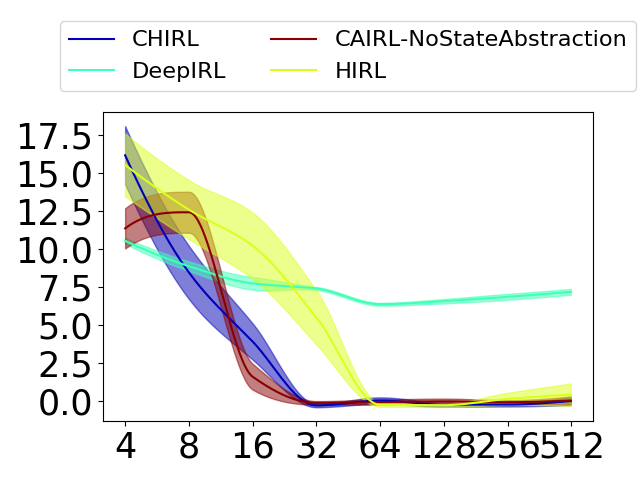}}
\hspace{3pt}
\subfloat[\TR]{\includegraphics[width=0.21\columnwidth]{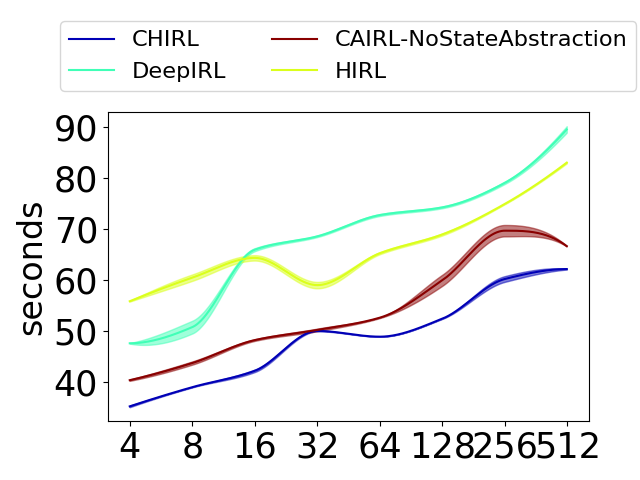}}

  \raisebox{-3\normalbaselineskip}[0pt][0pt]{\rotatebox[origin=c]{0}{\footnotesize (b)\;}}
\subfloat{\includegraphics[width=0.21\columnwidth]{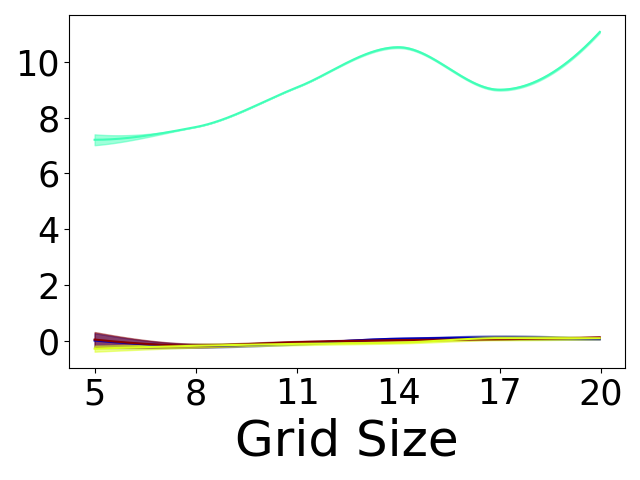}}
\hspace{5pt}
\subfloat{\includegraphics[width=0.21\columnwidth]{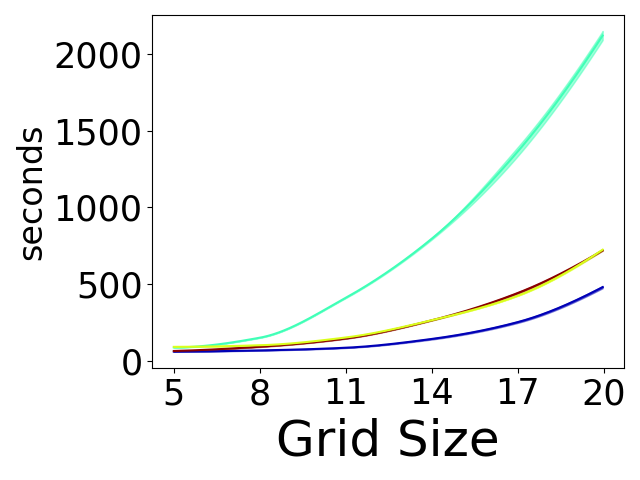}}
\caption{Performance benefits of data sharing and state abstraction across contexts for the \taxi domain. The solid line represents the mean and shaded area represents the standard deviation.}
\label{fig:ablation}
\end{wrapfigure}

Data sharing across contexts is 
one main benefit of our context hierarchy and is the essential difference between \algo and \cairlmh. 
Like \algo, \cairlmh  learns context-dependent rewards, but learns one for each context almost independently without data sharing. 
\figref{fig:ablation}
 shows that with sufficient data (at least $64$
trajectories), both \algo variants converge and learn good reward functions. However, without help from the context DAG and data sharing, \cairlmh takes much longer in training time and converges slower  than \algo. Further, a comparison of the two variants on all the three evaluation tasks (\figref{fig:env_quantitative}) indicate that the benefit of data sharing across contexts increases with the complexity of the task and the resulting context hierarchy.



\subsubsection{State Abstraction}
For the \taxi task, the context hierarchy contains
a MAXQ task graph consisting of three subtasks, and \algo learn the reward
of an HMDP.  
The MAXQ decomposition
allows for state abstraction by dropping state dimensions irrelevant for a
subtask. In particular, the passenger destination $d$ is not needed in
\textsc{Get}, the passenger's initial location is not needed in \textsc{Put},
and the passenger's location is not needed in \textsc{Nav}. Specifically, in  \textsc{Get},  \textsc{Put} and  \textsc{Nav}, the state space size reduces to $5 \times 5 \times 5 =125, 5 \times 5 \times 4 = 100, 5 \times 5 = 25$, respectively from the original state space size of  $5 \times 5 \times 5 \times 4 = 500$. 
For comparison, we created a
variant \cairlnsa, which removes the state abstraction, but retains the same reward function representation. \cairlnsa would then
have to solve an MDP with 500 states for all the \taxi subtasks. 
\figref{fig:ablation}\subfig a shows that both \algo and \algo-NoStateAbstraction learn good reward functions, but \algo is faster in training time with the help of state abstraction. 
To confirm, 
we performed additional experiments, which  increase the grid world size of \taxi  from $5 \times 5$ to $20 \times 20$. 
\figref{fig:ablation}\subfig b  shows that
 the training time gap between \algo and \cairlmh grows with increasing grid size. 
 Also, DeepIRL can not figure out the proper reward as before. 
 With the increasing size of the grid world, the training time of \algo grows slowly due to the state abstraction. 
 \algo solves an HMDP rather than a flat MDP. It makes \algo easily be adopted to large robotic tasks without the need of solving large MDPs with large state space and action space.

\section{Conclusion and Future Work}
\algo is a new IRL algorithm that exploits the hierarchically decomposed task
context to structure the reward function representation as a set of
interconnected neural network modules. The structured representation enables
data sharing across contexts and scales up IRL for learning complex reward
functions.  When a context hierarchy represents subtask decomposition, it is
closely related to the MAXQ task graph, which \algo exploits for hierarchical
MDP solution and benefits from the resulting subtask sharing and state
abstraction.
Experiments show promising results  on benchmark navigation tasks, including a large-scale
autonomous driving task in the CARLA simulator.

However, our current setup only considers tabular MDPs. We will extend to continuous space with motion planning in the future. Furthermore, we will explore more complex context-hierarchy to solve large realistic tasks, i.e. RoboCup. More interestingly, we want to try to connect \algo with TAMP~\cite{garrett2021integrated} architecture for a more generic solution for IRLs.  


\bibliographystyle{unsrt}
\bibliography{references}

\begin{thebibliography}{10}

\bibitem{NgRus00}
A.Y. Ng and S.J. Russell.
\newblock Algorithms for inverse reinforcement learning.
\newblock In {\em Proc. Int. Conf. on Machine Learning}, 2000.

\bibitem{ziebart2008maximum}
Brian~D Ziebart, Andrew~L Maas, J~Andrew Bagnell, and Anind~K Dey.
\newblock Maximum entropy inverse reinforcement learning.
\newblock In {\em AAAI}, volume~8, pages 1433--1438. Chicago, IL, USA, 2008.

\bibitem{wulfmeier2015maximum}
Markus Wulfmeier, Peter Ondruska, and Ingmar Posner.
\newblock Maximum entropy deep inverse reinforcement learning.
\newblock {\em arXiv preprint arXiv:1507.04888}, 2015.

\bibitem{finn2016guided}
Chelsea Finn, Sergey Levine, and Pieter Abbeel.
\newblock Guided cost learning: Deep inverse optimal control via policy
  optimization.
\newblock In {\em International Conference on Machine Learning}, pages 49--58,
  2016.

\bibitem{krishnan2016hirl}
Sanjay Krishnan, Animesh Garg, Richard Liaw, Lauren Miller, Florian~T Pokorny,
  and Ken Goldberg.
\newblock Hirl: Hierarchical inverse reinforcement learning for long-horizon
  tasks with delayed rewards.
\newblock {\em arXiv preprint arXiv:1604.06508}, 2016.

\bibitem{yu2019meta}
Lantao Yu, Tianhe Yu, Chelsea Finn, and Stefano Ermon.
\newblock Meta-inverse reinforcement learning with probabilistic context
  variables.
\newblock {\em arXiv preprint arXiv:1909.09314}, 2019.

\bibitem{dietterich2000hierarchical}
Thomas~G Dietterich.
\newblock Hierarchical reinforcement learning with the maxq value function
  decomposition.
\newblock {\em Journal of Artificial Intelligence Research}, 13:227--303, 2000.

\bibitem{garrett2021integrated}
Caelan~Reed Garrett, Rohan Chitnis, Rachel Holladay, Beomjoon Kim, Tom Silver,
  Leslie~Pack Kaelbling, and Tom{\'a}s Lozano-P{\'e}rez.
\newblock Integrated task and motion planning.
\newblock {\em Annual review of control, robotics, and autonomous systems},
  4:265--293, 2021.

\bibitem{bai2012online}
Aijun Bai, Feng Wu, and Xiaoping Chen.
\newblock Online planning for large mdps with maxq decomposition.
\newblock In {\em Proceedings of the 11th International Conference on
  Autonomous Agents and Multiagent Systems-Volume 3}, pages 1215--1216.
  International Foundation for Autonomous Agents and Multiagent Systems, 2012.

\bibitem{dosovitskiy2017carla}
Alexey Dosovitskiy, German Ros, Felipe Codevilla, Antonio Lopez, and Vladlen
  Koltun.
\newblock Carla: An open urban driving simulator.
\newblock {\em arXiv preprint arXiv:1711.03938}, 2017.

\bibitem{argall2009survey}
Brenna~D Argall, Sonia Chernova, Manuela Veloso, and Brett Browning.
\newblock A survey of robot learning from demonstration.
\newblock {\em Robotics and autonomous systems}, 57(5):469--483, 2009.

\bibitem{abbeel2004apprenticeship}
Pieter Abbeel and Andrew~Y Ng.
\newblock Apprenticeship learning via inverse reinforcement learning.
\newblock In {\em Proceedings of the twenty-first international conference on
  Machine learning}, page~1. ACM, 2004.

\bibitem{choi2013bayesian}
Jaedeug Choi and Kee-Eung Kim.
\newblock Bayesian nonparametric feature construction for inverse reinforcement
  learning.
\newblock In {\em Proc. Int. Jnt. Conf. on Artificial Intelligence}, pages
  1287--1293, 2013.

\bibitem{LevPop10}
S.~Levine, Z.~Popovic, and V.~Koltun.
\newblock Feature construction for inverse reinforcement learning.
\newblock In {\em Advances in Neural Information Processing Systems}, 2010.

\bibitem{levine2011nonlinear}
Sergey Levine, Zoran Popovic, and Vladlen Koltun.
\newblock Nonlinear inverse reinforcement learning with gaussian processes.
\newblock In {\em Advances in Neural Information Processing Systems}, pages
  19--27, 2011.

\bibitem{ziebart2010modeling}
Brian~D Ziebart, J~Andrew Bagnell, and Anind~K Dey.
\newblock Modeling interaction via the principle of maximum causal entropy.
\newblock 2010.

\bibitem{rothkopf2013modular}
Constantin~A Rothkopf and Dana~H Ballard.
\newblock Modular inverse reinforcement learning for visuomotor behavior.
\newblock {\em Biological cybernetics}, 107(4):477--490, 2013.

\bibitem{andre2002state}
David Andre and Stuart~J Russell.
\newblock State abstraction for programmable reinforcement learning agents.
\newblock In {\em AAAI/IAAI}, pages 119--125, 2002.

\bibitem{FraHo17}
K.~Frans, J.~Ho, X.~Chen, P.~Abbeel, and J.~Schulman.
\newblock Meta learning shared hierarchies.
\newblock In {\em Proc. Int. Conf. on Learning Representations}, 2017.

\bibitem{Kor85}
R.E. Korf.
\newblock Macro-operators: A weak method for learning.
\newblock {\em Artificial Intelligence}, 26(1):35--77, 1985.

\bibitem{Kno90}
C.A. Knoblock.
\newblock Learning abstraction hierarchies for problem solving.
\newblock In {\em Proc. AAAI Conf. on Artificial Intelligence}, 1990.

\bibitem{sutton1999between}
Richard~S Sutton, Doina Precup, and Satinder Singh.
\newblock Between mdps and semi-mdps: A framework for temporal abstraction in
  reinforcement learning.
\newblock {\em Artificial intelligence}, 112(1-2):181--211, 1999.

\bibitem{hwang2019option}
Rakhoon Hwang, Hanjin Lee, and Hyung~Ju Hwang.
\newblock Option compatible reward inverse reinforcement learning.
\newblock {\em arXiv preprint arXiv:1911.02723}, 2019.

\bibitem{venuto2020oirl}
David Venuto, Jhelum Chakravorty, Leonard Boussioux, Junhao Wang, Gavin
  McCracken, and Doina Precup.
\newblock oirl: Robust adversarial inverse reinforcement learning with
  temporally extended actions.
\newblock {\em arXiv preprint arXiv:2002.09043}, 2020.

\bibitem{qureshi2018adversarial}
Ahmed~H Qureshi, Byron Boots, and Michael~C Yip.
\newblock Adversarial imitation via variational inverse reinforcement learning.
\newblock {\em arXiv preprint arXiv:1809.06404}, 2018.

\bibitem{chatterjee2008value}
Krishnendu Chatterjee and Thomas~A Henzinger.
\newblock Value iteration.
\newblock In {\em 25 Years of Model Checking}, pages 107--138. Springer, 2008.

\bibitem{syed2008game}
Umar Syed and Robert~E Schapire.
\newblock A game-theoretic approach to apprenticeship learning.
\newblock In {\em Advances in neural information processing systems}, pages
  1449--1456, 2008.

\bibitem{pirotta2016inverse}
Matteo Pirotta and Marcello Restelli.
\newblock Inverse reinforcement learning through policy gradient minimization.
\newblock In {\em Proceedings of the AAAI Conference on Artificial
  Intelligence}, volume~30, 2016.

\bibitem{brown2018efficient}
Daniel Brown and Scott Niekum.
\newblock Efficient probabilistic performance bounds for inverse reinforcement
  learning.
\newblock In {\em Proceedings of the AAAI Conference on Artificial
  Intelligence}, volume~32, 2018.

\bibitem{rathi2019multi}
Ghanshyam Rathi, Hilda Faraji, Nikhil Gupta, Christian Traub, Michael
  Schaffner, and Goerg Pflug.
\newblock Multi-camera dynamic top view vision system, January~15 2019.
\newblock US Patent 10,179,543.

\bibitem{wulfmeier2017large}
Markus Wulfmeier, Dushyant Rao, Dominic~Zeng Wang, Peter Ondruska, and Ingmar
  Posner.
\newblock Large-scale cost function learning for path planning using deep
  inverse reinforcement learning.
\newblock {\em The International Journal of Robotics Research},
  36(10):1073--1087, 2017.

\bibitem{romera2018erfnet}
Eduardo Romera, Jos{\'e}~M Alvarez, Luis~M Bergasa, and Roberto Arroyo.
\newblock Erfnet: Efficient residual factorized convnet for real-time semantic
  segmentation.
\newblock {\em IEEE Transactions on Intelligent Transportation Systems},
  19(1):263--272, 2018.

\end{thebibliography}

\clearpage
\appendix
\section*{Proof of MaxEnt Gradient}
The complete negative log likelihood of data given reward $\vec{R}$ is~\cite{levine2011nonlinear}:
\begin{align}
\mathcal{L_D} 
&= -\log P(\mathcal{D}|\vec{R}) \\
&= -\sum_{i,t} \log P(a_{i,t}|s_{i,t}) \\ 
&= -\sum_{i,t} (\vec{Q}_{s_{i,t}, a_{i,t}}^{\vec{R}} - \vec{V}_{s_{i,t}}^{\vec{R}}) \\
&= -\sum_{i,t} (\vec{R}_{s_{i,t}} + \sum_{s'}\gamma T(s_{i,t}, a_{i,t}, s')\vec{V}_{s'}^{\vec{R}} - \vec{V}_{s_{i,t}}^\vec{R})
\end{align}

where the value function $\vec{V}_{s}^{\vec{R}}$ is a "soft" version of Bellman update: $\vec{V}_{s}^{\vec{R}} = \log \sum_a \exp \vec{Q}_{sa}^{\vec{R}}$.

Let's define $\hat{\mu}^a_D$ as state-action visitation frequency of the expert demonstrations, where $\hat{\mu}^a_D(s) = \sum_{i,t} 1_{s_{i,t} = s \land a_{i,t} = a}$. $\hat{\mu}_D = \sum_a \hat{\mu}^a_D$ is the state visitation frequency, $E[\mu \vert s] = \frac{\partial \vec{V}_s^{\vec{R}}}{\partial \vec{R}}$ indicates the expected state visitation count when starting from state s and following the optimal stochastic policy.  
Differentiating with respect to the reward, we obtain the following gradient:
{\scriptsize 
\begin{align}
\frac{\partial \mathcal{L_D}}{\partial \vec{R}}  
& =  -\sum_{i,t}\frac{\partial \vec{R}_{s_{i,t}}}{\partial \vec{R}} + \sum_{i,t} (\sum_{s'}\gamma T(s_{i,t}, a_{i,t}, s')\frac{\partial \vec{V}_{s'}^{\vec{R}}}{\partial \vec{R}} - \frac{\partial \vec{V}_{s_{i,t}}^{\vec{R}}}{\partial \vec{R}}) \\
& =  -\sum_{i,t}\frac{\partial \vec{R}_{s_{i,t}}}{\partial \vec{R}} - \sum_{i,t} (\frac{\partial \vec{V}_{s_{i,t}}^{\vec{R}}}{\partial \vec{R}} - \sum_{s'}\gamma T(s_{i,t}, a_{i,t}, s')\frac{\partial \vec{V}_{s'}^{\vec{R}}}{\partial \vec{R}}) \\
&= -(\hat{\mu}_D - \sum_{i,t} (E[\mu \vert s_{i,t}] - \sum_{s'}\gamma T(s_{i,t}, a_{i,t}, s')E[\mu \vert s']))  \\
&= -\hat{\mu}_D + \sum_{i,t}(\sum_{a,s} E[\mu \vert s_{i,t}, a, s] - \sum_{s'}\gamma T(s_{i,t}, a_{i,t}, s')E[\mu \vert s'] \\
&= -\hat{\mu}_D + \sum_{i,t}(\sum_{a,s} 1_{s_{i,t}=s\land a_{i,t}=a} E[\mu \vert s_{i,t}, a_{i,t}, s] \\&- 
 \sum_{s'}\gamma T(s_{i,t}, a_{i,t}, s')E[\mu \vert s']) \\
&= -\hat{\mu}_D + \sum_{a,s} \hat{\mu}^a_D(s) E[\mu \vert s] - \sum_{i,t, s'}\gamma T(s_{i,t}, a_{i,t}, s')E[\mu \vert s'] \\
&= -\hat{\mu}_D + \sum_{s} \hat{\mu}_D(s) E[\mu \vert s] - \sum_{i,t, s'}\gamma T(s_{i,t}, a_{i,t}, s')E[\mu \vert s'] \\
&= -\hat{\mu}_D + \sum_{s} E[\mu \vert s](\sum_a \hat{\mu}_D(sa) - \sum_{i,t}\gamma T(s_{i,t}, a_{i,t}, s)) 
\end{align}
}
Let's define $\hat{\nu}_D$ as the initial state distribution, where $\hat{\nu}_D(s) = \sum_a \hat{\mu}^a_D(s) - \sum_{i, t} \gamma T(s_{i,t}, a_{i,t}, s)$.
Then we can have:
\begin{align}
\frac{\partial \mathcal{L_D}}{\partial \vec{R}}  
&= -\hat{\mu}_D + \sum_s \hat{\nu}_D(s)E(\mu\vert s) \\
&= -\hat{\mu}_D + \hat{\nu}_DE(\mu)  \\
&= \hat{\mu}_\vec{R} - \hat{\mu}_D
\end{align}

\section*{Experiment HyperParameters}
We run all the experiments multiple runs on a server with  Nvidia GeForce GTX 2080 Ti.~\tabref{app:param} shows the exact hyperparameters for each benchmark. For all the hyperparameters used in the experiments, \ie learning rate, L1 norm, and L2 norm \etc, we performed random search and grid search to find the proper hyperparameters for each experiment. Random search is firstly used to guess the lower bound and upper bound of these hyperparameters. Another round of grid search, which uniformly chopped the random search intervals into a set of fixed intervals, is used to find the best hyperparameters. We observe that the choice of the L1 norm and L2 norm affects the performance a lot during the hyperparameter search. For a fair comparison, the best hyperparameters were used for each method.

\begin{table}[ht]
\caption{Benchmark hyperparameters}
\centering
\label{app:param}
\fontsize{6.5}{9}
\selectfont
\begin{tabular}{|c|c|c|c|c|}
\hline
Benchmarks        & Learning Rate & L1 Norm & L2 Norm & Number of Epochs \\ \hline
\goalnav    & 1e-2          & 1e-3    & 8e-1    & 500              \\ \hline
\intnav & 5e-2          & 2       & 8e-1    & 500              \\ \hline
\taxi        & 1e-2          & 1e-3    & 8e-1    & 500              \\ \hline
CARLA        & 1e-3          & 1e-3    & 8e-1    & 500              \\ \hline
\end{tabular}
\end{table}

\section*{TaxiDomain}

\begin{wrapfigure}{r}{0.5\textwidth} 
  \center
  \begin{tabular}{c@{\hspace{0.5in}}c}
    \includegraphics[height=0.8in]{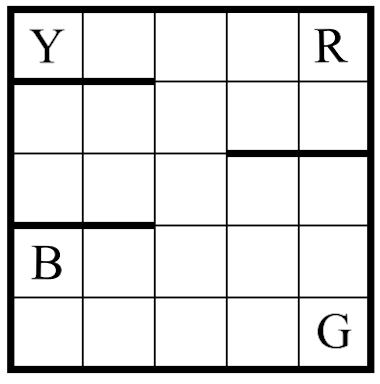}
    &\includegraphics[height=0.8in]{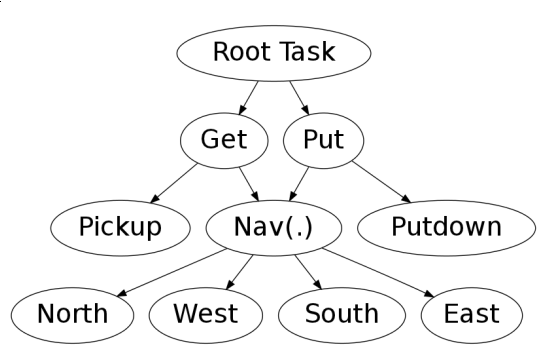}     \\
    (\subfig a) & (\subfig b)
  \end{tabular}
  \caption{(\subfig a) Taxi Domain task. (\subfig b) MAXQ Task Graph for taxi task.}
  \label{fig:taxi_graph}
\end{wrapfigure}

\label{app:taxidomain}
This section presents the details of TaxiDomain.~\figref{fig:taxi_graph} shows the taxi environment and corresponding task graph decomposition. Here, we present how to solve the original task with \algo when the complex task graph prior knowledge is accessable. 
Note that although there is state abstraction in the leaf node MDPs, the raw input state is still a four tuple $(x,y,pl,dl)$ to be consistent to share features for all leaf nodes. For those leaf nodes with state abstraction, the unused dimension is marked as $IGNORE$. For example, in $Get$, the passenger's destination is not considered, then the state tuple would be $(x, y, pl, IGNORE)$, any $dl$ is transferred to $IGNORE$ in this dimension. Thus it can be shortened as $(x,y,pl)$. 

Formally, in $Get$ leaf node, the state is three tuple $s = (x, y, pl)$, the termination predicate is:
\[
\begin{array}{lcl}
  \tiny
  T(s, Nav(c)) &=& (x_c, y_c, pl) \\
  T(s, Pickup) &=&
\begin{cases}
(x,y,t), & \text{if} \, (x,y) == (pl.x, pl.y) \\
(x,y,pl), & \text{otherwise}
\end{cases}
\end{array}
\]
where $(x_c, y_c)$ is the position of terminal $c$ and $t$ indicates the passenger is in the taxi.

Similarly, in $Put$ leaf node, the state space contains $5 \times 5 \times 4 = 100$ states. There are four macro actions, $NavR, NavG, NavY, NavB$ and one primitive action $Putdown$.

In $Put$ leaf node, the state is three tuple $s = (x, y, dl)$, the termination predicate is:
\[
\begin{array}{lcl}
  T(s, Nav(c)) &=& (x_c, y_c, dl) \\
  T(s, Putdown) &=&
\begin{cases}
  (x,y,pl), & \text{if} \, (x, y, dl) == (pl.x, pl.y,t) \\
  (x,y,t), & \text{otherwise}
\end{cases}
\end{array}
\]
In the $Nav(c)$ leaf node, the state space contains $5 \times 5 = 25$ states. The action space only contains four primitive actions: $N, S, W, E$. It is a normal MDP the same as that in the GoalNavigation benchmark.

\textbf{Execution of the learned policy} After learned the leaf node rewards, we can induce the policy of different levels. Executing the learned policy is similar to the subroutine-call-return schema. We will execute the action only when it is a primitive action, otherwise, we just call the corresponding subtask for macro actions. For example, delivering any passenger would call $Get$ and $Put$. To finish $Get$, the policy will give action on the current state. It can be either $Pickup$ or macro action $NavR, NavG, NavR, NavB$. If a macro is called, it will query the subtask policy. \ie $NavR$ will give a sequence of primitive actions of $N, S, W, E$ to the final $R$ terminal. After $Pickup$ action is executed, $Put$ subtask would be executed by firstly navigating to the destination via calling $Nav(c)$ and $Putdown$ the passenger to finish the whole task.

\section*{Network Architecture For Carla Experiments}
\begin{figure}
\vspace{-15pt}
\centering
\captionsetup[subfigure]{font=scriptsize,labelfont=scriptsize,position=top,labelformat=empty}
  \raisebox{-3\normalbaselineskip}[0pt][0pt]{\rotatebox[origin=c]{0}{\footnotesize (a)\;}}
\subfloat{\includegraphics[width=0.29\columnwidth]{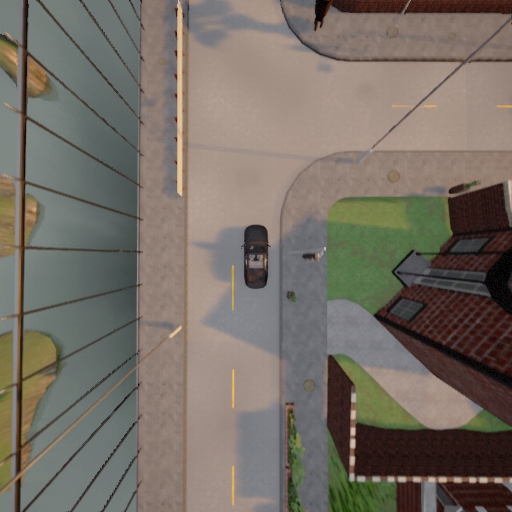}}
\hspace{0.02\columnwidth}
\subfloat{\includegraphics[width=0.29\columnwidth]{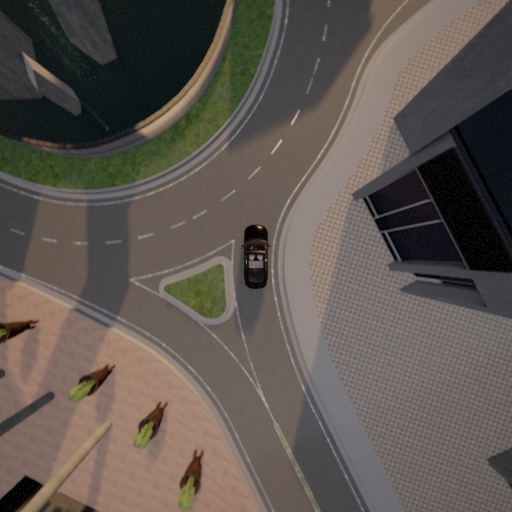}}
\hspace{0.02\columnwidth}
\subfloat{\includegraphics[width=0.29\columnwidth]{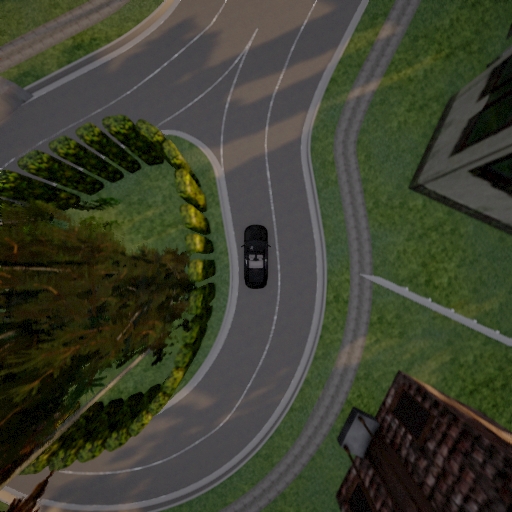}}

  \raisebox{-3\normalbaselineskip}[0pt][0pt]{\rotatebox[origin=c]{0}{\footnotesize (b)\;}}
\subfloat{\includegraphics[width=0.29\columnwidth]{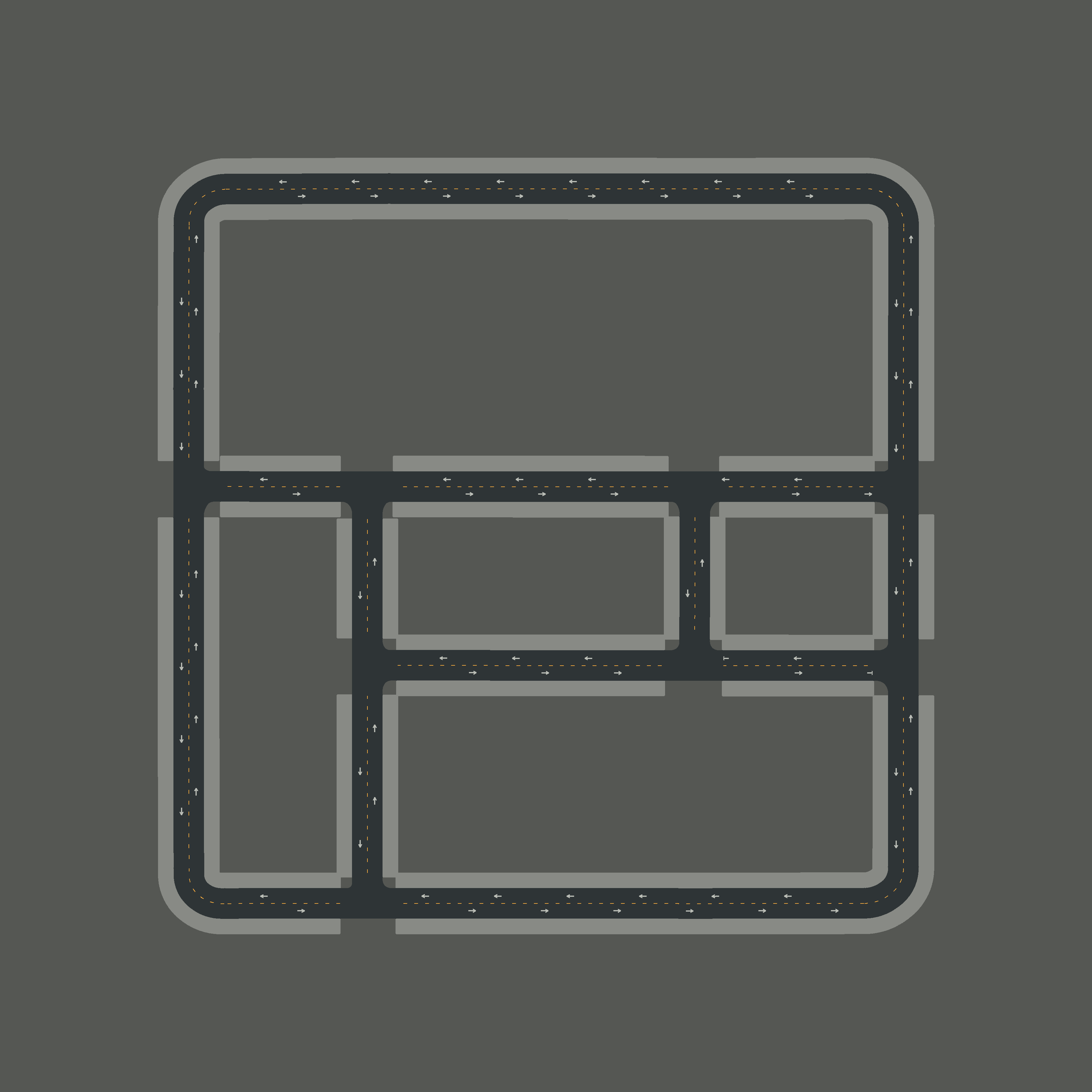}}
\hspace{0.02\columnwidth}
\subfloat{\includegraphics[width=0.29\columnwidth]{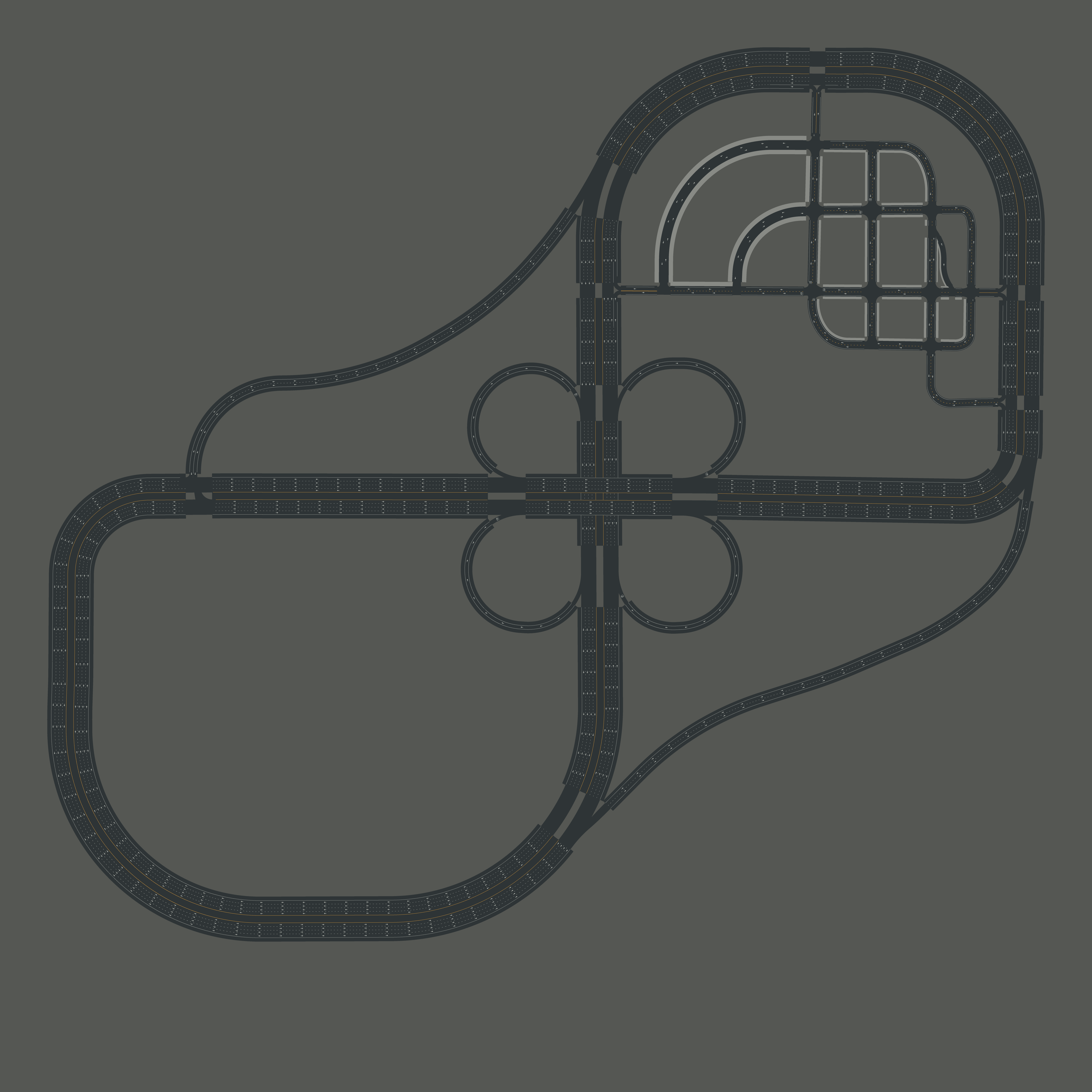}}
\hspace{0.02\columnwidth}
\subfloat{\includegraphics[width=0.29\columnwidth]{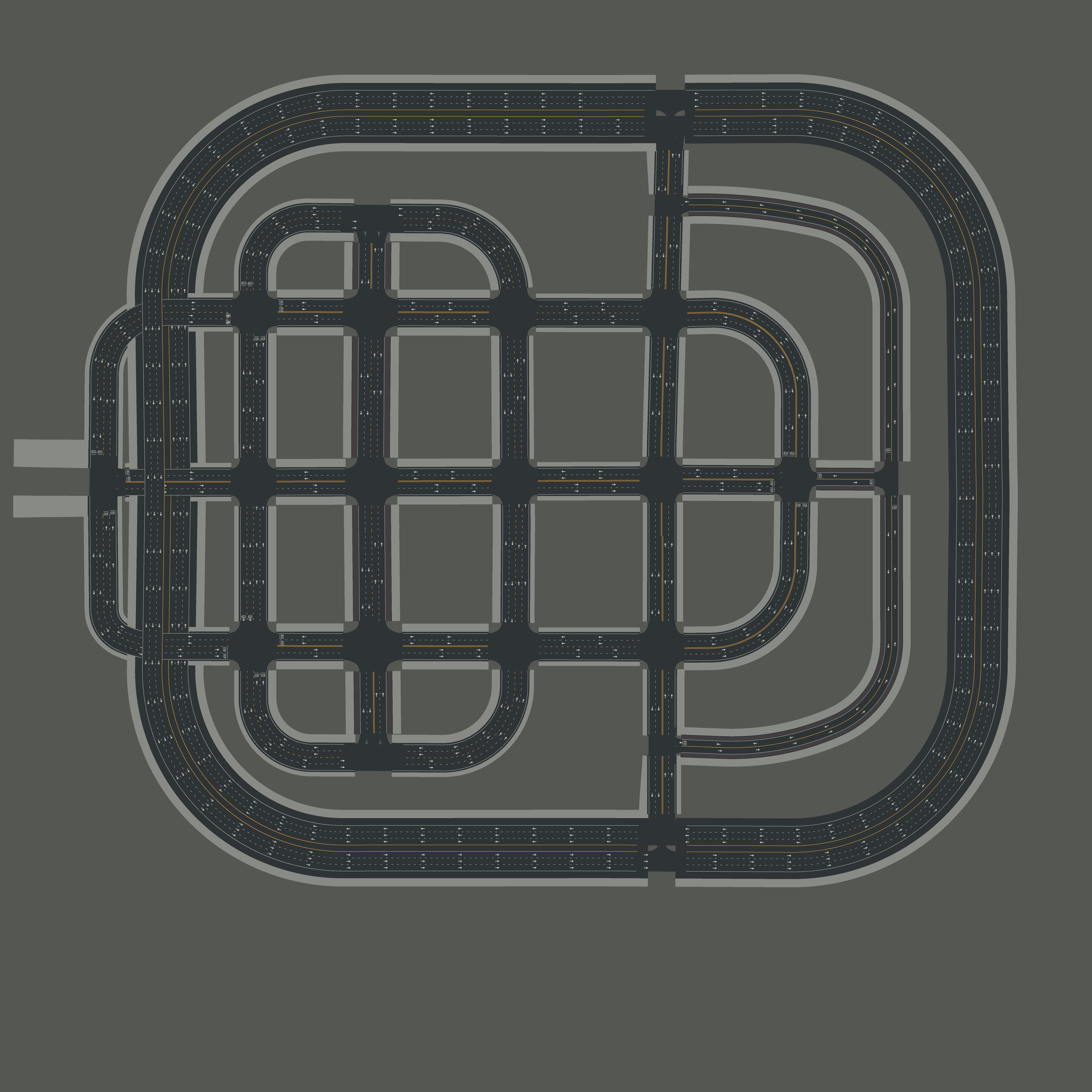}}
\caption{Carla Simulator setup. \subfig{(a)} Sample top-down images from
  given towns. \subfig{(b)} Different town maps used for data collection in
  the CARLA Simulator.}
\label{fig:carla_setup}
\end{figure}

For the large scale CARLA experiment, we collect data in CARLA~(\figref{fig:carla_setup}) simulation in various towns and paths. For this example, we encode the traffic rules and regional rules as context hierarchy. It is intuitive to encode any semantic information in our \algo framework. One big benefit of \algo is that we can use the explicit context information to control the calling of subtasks, the executor has the full control of all the behaviors. For example, it is also very easy to encode traffic lights into our framework. If the traffic light is red, we can easily parse the DAG to reach the stop behavior while pass-through behavior with green traffic light. 

\begin{figure}
\centering
\includegraphics[width=0.9\columnwidth]{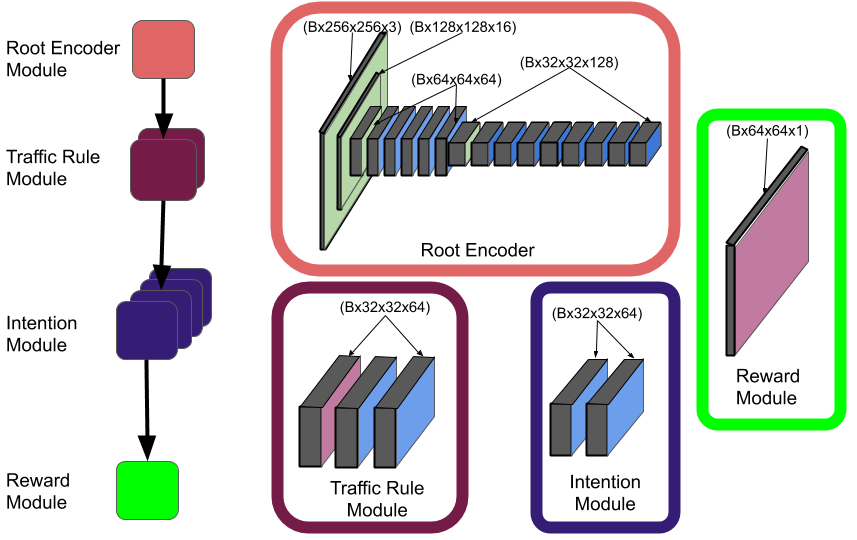}
\caption{Neural network architecture for cost function learning
  with image input.
  All DownsampleBlock is plotted with color light green, UpsampleBlock with
  light red and NonBottleneck1d with light blue. Each slice of context
  variable module are plotted in details on the right.
}
\label{fig:arch}
\end{figure}

\label{sec:na}
The neural network architecture is built from three basic modules~(\figref{fig:arch}):
DownsampleBlock, UpsampleBlock and NonBottleneck1d. DownsampleBlock is the core module used in the Encoder to extract features from high-dimensional image inputs to feature maps. UpsampleBlock is used to upsample the feature maps to reconstruct the corresponding image segment. NonBottleneck1d is the efficient factorized residual blocks as the residual layers for deep neural networks. They are the same building blocks from ~\cite{romera2018erfnet}. Each DownsampleBlock has a conv2d layer with $3 \times 3$ kernels, a max-pool layer and a batch norm layer. Each NonBottleneck1d is the residual factorized block with two $1 \times 3$ and two $3 \times 1 $ 1D filters and a direct identity mapping. The whole architecture is plotted in~\figref{fig:arch}. The input of the neural networks is the ego-centric bird view image. It is re-scaled to the size of 256 by 256. A batch of images forms a tensor of $B \times 256 \times 256 \times 3$, where $B$ is the number of images in a batch. The output is the pixel level cost map function with the size $B \times 64 \times 64$, where 64 by 64 is the MDP grid world size. All the context variable modules are constructed by these three basic blocks.  

\begin{figure}
\hspace{-15pt}
  \centering
  \begin{tabular}{cccc}
  &{\small (a) LeftSidedDrive} &&{\small (b) RightSidedDrive} \\
      \vspace{3pt}
            \tiny LF &   
      \includegraphics[align=c, width=0.25\columnwidth]{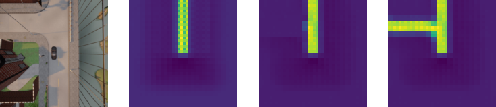}&    
            \tiny LF       & 
      \includegraphics[align=c, width=0.25\columnwidth]{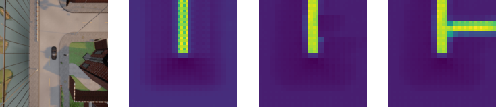}\\
      \vspace{3pt}
  
            \tiny LF &   
      \includegraphics[align=c, width=0.25\columnwidth]{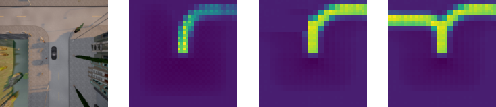}&    
            \tiny LF       & 
      \includegraphics[align=c, width=0.25\columnwidth]{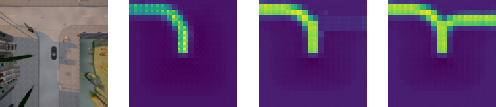}\\
  
  
      \vspace{3pt}
            \tiny TL &   
      \includegraphics[align=c, width=0.25\columnwidth]{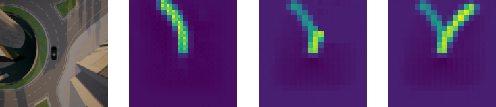}&    
            \tiny TR       & 
      \includegraphics[align=c, width=0.25\columnwidth]{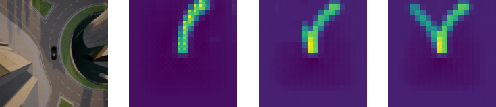}\\
  
  
      \vspace{3pt}
            \tiny TL &   
      \includegraphics[align=c, width=0.25\columnwidth]{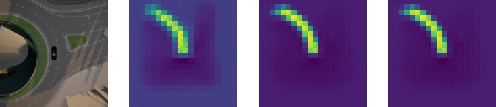}&    
            \tiny TR       & 
      \includegraphics[align=c, width=0.25\columnwidth]{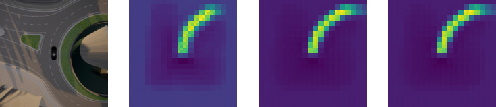}\\
  
      \vspace{3pt}
            \tiny TR       & 
      \includegraphics[align=c, width=0.25\columnwidth]{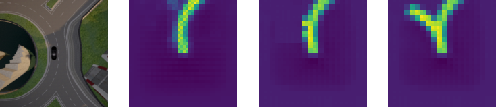}&
            \tiny TL &   
      \includegraphics[align=c, width=0.25\columnwidth]{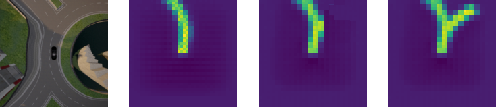}\\    
  
      \vspace{3pt}
            \tiny TR &   
      \includegraphics[align=c, width=0.25\columnwidth]{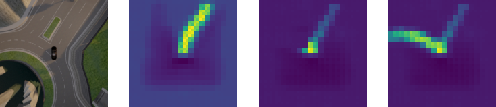}&    
            \tiny TL       & 
      \includegraphics[align=c, width=0.25\columnwidth]{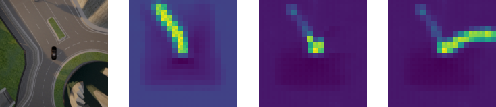}\\
  
      \vspace{3pt}
            \tiny ST &   
      \includegraphics[align=c, width=0.25\columnwidth]{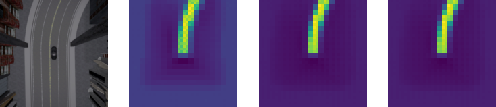}&    
            \tiny ST       & 
      \includegraphics[align=c, width=0.25\columnwidth]{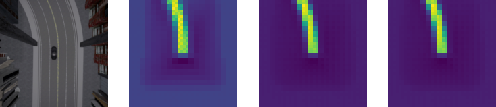}\\
      
            \vspace{3pt}
            \tiny LF &   
      \includegraphics[align=c, width=0.25\columnwidth]{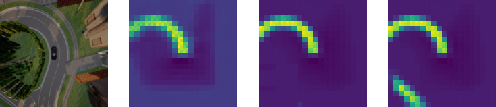}&    
            \tiny LF       & 
      \includegraphics[align=c, width=0.25\columnwidth]{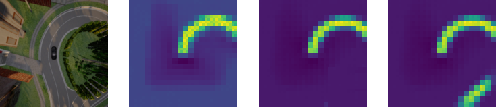}\\
  \end{tabular}
  \caption{Visualization of the learned rewards.The brighter place has higher reward (lower cost) for driving. Each row shows two different learned reward with different first level contexts(\subfig a) LeftSidedDrive and (\subfig b) RightSidedDrive. Different rows show different performances with different second level contexts: Turning Left (TL), Turning Right(TR), Going Straight (ST) and Lane Following (LF). Four tuple of shown images are image observations, \algo reward, \cairlmh reward and deepIRL reward.}
  \label{fig:carla}
  \hspace{15pt}
  \end{figure}

\figref{fig:carla} shows the learned underlying reward in CARLA environment. The induced policy performance from the learned reward are summarized in~\tabref{tab:table1}. We can easily see the benefits of explicit context-hierarchy. Also, \algo makes use of the powerful feature representation learning of neural network so that it can recover complex rewards give high-dimensional image input. The collected data for this large scale training can be downloaded from \href{https://drive.google.com/file/d/18jub_iZBvhlebX2GQux3tTcdMHqjh-WK/view?usp=sharing}{\textcolor{blue}{google drive}}.

\end{document}